\documentclass[a4paper]{article}

\usepackage{INTERSPEECH_v2}

\title{Experiments on Open-Set Speaker Identification with Discriminatively Trained Neural Networks}
\name{Stefano Imoscopi \sthanks{~~Research conducted as an intern at Ericsson Research.}$^\ast$, Volodya Grancharov$^\dagger$, Sigurdur Sverrisson$^\dagger$, Erlendur Karlsson$^\dagger$, \\ Harald Pobloth$^\dagger$}
\address{
  $^\ast$School of Electrical Engineering and Computer Science, KTH, 100 44, Stockholm, Sweden\\
  $^\dagger$Ericsson Research, Ericsson AB, 164 80 Stockholm, Sweden}
\email{stefano.imoscopi@gmail.com, \{volodya.grancharov, sigurdur.sverrisson, erlendur.karlsson, harald.pobloth\}@ericsson.com}

\begin{document}

\maketitle
\begin{abstract}
This paper presents a study on discriminative artificial neural network classifiers in the context of open-set speaker identification. Both $2$-class and multi-class architectures are tested against the conventional Gaussian mixture model based classifier on enrolled speaker sets of different sizes. The performance evaluation shows that the multi-class neural network system has superior performance for large population sizes.
\end{abstract}


\section{Introduction} \label{sec:intro}
In general \textit{automatic speaker recognition} refers to computer algorithms that can recognize persons (persons identity) from samples of their voice. Based on the exact definition, there are three major classes of speaker recognition tasks. First, and the most researched class is \textit{speaker verification}, where an identity is claimed by the user, and a binary decision is made whether to accept or reject this claim. Second class is \textit{closed-set speaker identification}, which refers to the problem of labeling audio sample as belonging to one of a set of known voices. Third class, \textit{open-set speaker identification}, is a natural extension of the second class with the possibility that the speaker might be outside of the set of known voices (set of enrolled speakers). This paper deals exclusively with the third class speaker recognition problem. However, we first give a short overview of the relevant technology advances, often developed in the context of verification or closed-set identification tasks.

For many years the dominant approach for text-independent speech recognition was based on Gaussian Mixture Models (GMMs) \cite{reynolds1995robust, reynolds2000speaker}. In this type of systems, a generative type of learning is employed, where probability distribution of feature vectors, associated with a particular speaker, is modelled as a mixture of Gaussian distributions. 
Discriminative type of classifiers, such as Support Vector Machine (SVM) have also been studied in the context of speaker recognition task \cite{campbell2006support, campbell2003phonetic}. More recent adavances in the field are related to the introduction of Gaussian supervectors in combination with SVM \cite{campbell2006svm}, as well as Joint Factor Analysis (JFA) based methods \cite{kenny2005joint}. Combination of these two algorithms results in the i-vector concept \cite{dehak2011front}, currently considered state-of-the-art in the area of speaker recognition.

Experiments with Artificial Neural Networks (NNs) in the field of speaker recognition can be dated back to the 1990s, however a recent spike of interest was triggered by the achievements obtained using NNs for Automatic Speech Recognition (ASR) \cite{hinton2012_ASR_DNN, Dahl2012_ASR_DNN}. Majority of the studies concerned with NN in the area of speaker recognition, were related to either using NN in the role of robust, non-linear feature extractor, or combining output from different classifiers. For example NN were used as bottleneck features extraction in \cite{yamada2013bottleneckDNN} or to extract Baum-Welch statistics for i-vector-based sysem, e.g., \cite{kenny2014DNN_Baum_Welch}. Examples of NNs used to combine the scores of multiple classifiers can be found in \cite{navratil2001} and \cite{xiang2003}. In a role of classifiers, discriminatively trained multi-class NNs ("all-together" training configuration) have been used in closed-set identification scenario in \cite{bennani1991NN, Hossain2007}. Historically, $2$-class NNs ("one-against-all" or "one-against-one" training configuration) were attracting attention mainly because of their computational tractability and scalability advantages. Such systems were studied in \cite{oglesby1990firstNN, rudasi1991one-against-one} in the scenario of closed-set identification, then extended also to the verification task in \cite{wang1997GammaNN} and more recently in \cite{ghahabi2014ivect2NN, safari2015}.

The focus on this study is on the text-independent open-set speaker identification problem, which is of critical importance to multimedia annotation applications, like indexing news, film archives, and in general, historical audio documents. Open-set identification algorithms have also applications in audio database search task for recorded meetings and telephone interactions. The particular scenario considered throughout the presentation is search in audio archives for a set of enrolled speakers. This influences the design choices, like using large enrolled and impostor sets, wideband microphone speech, etc. Design choices and system configuration are discussed in more details in the next sections. Despite the fact that the open-set identification is the most challenging of speaker recognition problems, the research effort has been quite limited, in comparison to the speaker verification task. Generative GMM approach was studied in \cite{do211open_set, zigel2006open_set} and experiments with i-vector reported in \cite{karadaghi2014open_set}. Attempts to use discriminative classifiers can be found in \cite{brew2009open_set}, where SVM was applied to combine Universal Background Model (UBM) and cohort normalization, and in \cite{gao211open_set}, where discriminative GMM training was tested. There are no conclusive reports on the applicability of NN classifiers in the context of open-set identification. Use of $2$-class NNs was discussed in \cite{ganchev2002open_set}, with focus on training with limited material (only 128 feature vectors per-speaker). In the classical paper \cite{farrel1994NN_and_convClass}, among other classification techniques $2$-class NNs were also studied. However, due to the limited computational power available at that time only 20 enrolled speakers and 18 impostors were used in simulations. In this paper we made an attempt to characterize the performance and other practical aspects of the large, discriminatively trained NN classifiers in the context of open-set speaker identification.

\section{Considered classification architectures} \label{sec:Architecture}
To study the performance of generative vs. discriminative trained open-set speaker identification systems, we consider three classification architectures. The first one is the conventional GMM based classifier, which serves as a benchmark system. It is a generative model that estimates the feature distribution for each speaker and it is described in more details in Section~\ref{sec:GMM}. The second one is a $2$-class NN, and the third one is a multi-class NN system. For these two systems a multilayer feedforward neural network with fully connected architecture \cite{Duda:01} is considered. These are discriminative models, trained to model boundaries between different classes, with the differences being that $2$-class NN performs "one-against-all" training, thus learning to discriminate between a particular speaker and the class of "general population", while multi-class NN performs "all-together" training and learns to discriminate between large number of speakers. More detailed descriptions are provided in Section~\ref{sec:subNN} and Section~\ref{sec:multi-NN}.

From an architectural point of view the goal of open-set identification can be achieved in a two step process: 1) \textit{Closed-Set Identification Step}: identifying the speaker model in the enrolled set, which best matches the test utterance and 2) \textit{Verification Step}, taking decision whether the test utterance has actually been produced by the speaker associated with the best-matched model, or by unknown speaker outside the enrolled speaker set. These two steps are realized differently in the three systems under investigation, but in all cases the two steps are highlighted to ease the comparison.

We introduce some notation, to facilitate discussion on different classification architecture presented below. The set of enrolled speakers is denoted as $\{s_k \}_{k=1}^{K}$, where $K$ is the population size. At the recognition state, we denote the total set of feature vectors from an audio utterance, sent for classification, as $X$. This feature set consists of all feature vectors $x_n$, extracted form a short-time audio segment: ${X} = \{x_n \}_{n=1}^{N}$, where $N$ is the number of those segments.

\subsection{Generative GMM architecture} \label{sec:GMM}
\begin{figure}[!thb]
	\centering
	\includegraphics[width=0.45\textwidth]{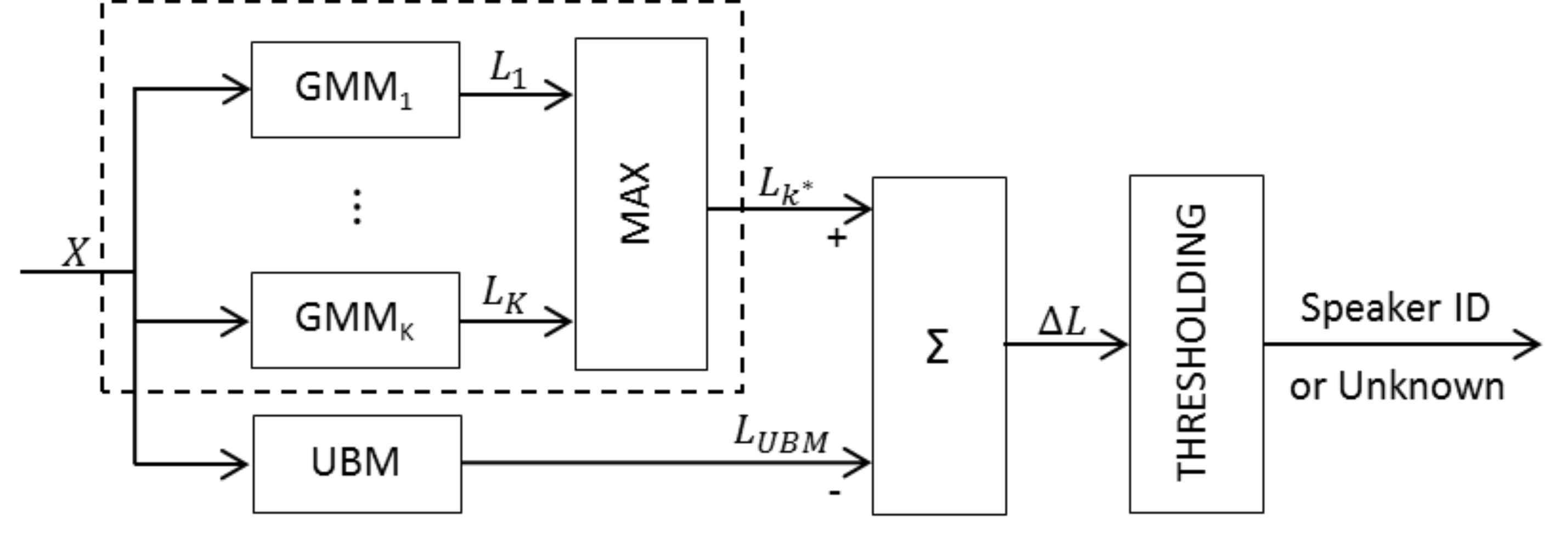}
	\caption{Conventional GMM architecture with UBM score normalization for open-set speaker identification scenario.}
	\label{fig:GMM-Architecture}
\end{figure}

The way a conventional GMM classifier is used in an open-set identification task is illustrated in Figure \ref{fig:GMM-Architecture}. The input to the open-set identification system is a feature vector set $X$ extracted from an audio recording associated with the speaker under test. The outcome of such system is either the speaker identity (ID from the enrolled list) or a decision that the speaker is unknown.

\textit{Closed-Set Identification Step} is solved by a maximum likelihood classifier, which corresponds to the "dashed box" in Figure \ref{fig:GMM-Architecture}. GMM based classifier operates on a feature vector-by-vector basis. Given a specific feature vector $x_n$, each speaker model $\omega_{k}$ associates a number corresponding to the degree of match with that speaker (likelihood that the data is generated by this model). Let $k^{*}$ be the index of the most likely speaker, then the output of this \textit{Closed-Set Identification} module is the mean log-likelihood of $X$ given the GMM speaker model $\omega_{k^{*}}$:
\begin{equation} \label{eq:L_Best}
	L_{k^{*}}\left( X \right) = \frac{1}{N} \sum_{n=1}^{N} \log p\left(x_n | \omega_{k^{*}} \right).
\end{equation}

In the \textit{Verification Step} the obtained $L_{k^{*}}$ could not be thresholded directly, instead a Likelihood Ratio (LR) approach \cite{neyman1933LR} is used. An additional model, UBM \cite{reynolds1995UBM} is used to normalize the likelihood score of the most likely speaker model. UBM models the distribution of the feature space of the "average" speaker, and it is created by pooling features from a large number of speakers and training a single, large GMM. Similarly to the equation~\ref{eq:L_Best}, normalization score $L_{UBM}$ is defined as the mean log-likelihood for the same feature set $X$ given $\omega_{UBM}$, the normalized score is calculated as:
\begin{equation} \label{eq:LR}
\Delta L\left(X\right) = L_{k^{*}}\left(X\right) - L_{UBM}\left(X\right),
\end{equation}
and the decision is based on comparing it to a speaker-independent threshold $\theta$:
\begin{equation} \label{eq:ver1}
\Delta L\left( X \right) \begin{cases} 
  \geq\theta, & X \text{ is spoken by } s_{k^{*}} \\ 
  <\theta, & X \text{ is not spoken by } s_{k^{*}} 
  \end{cases}
\end{equation}

\subsection{$2$-class NN architecture} \label{sec:subNN}
\begin{figure}[!thb]
	\centering
	\includegraphics[width=0.45\textwidth]{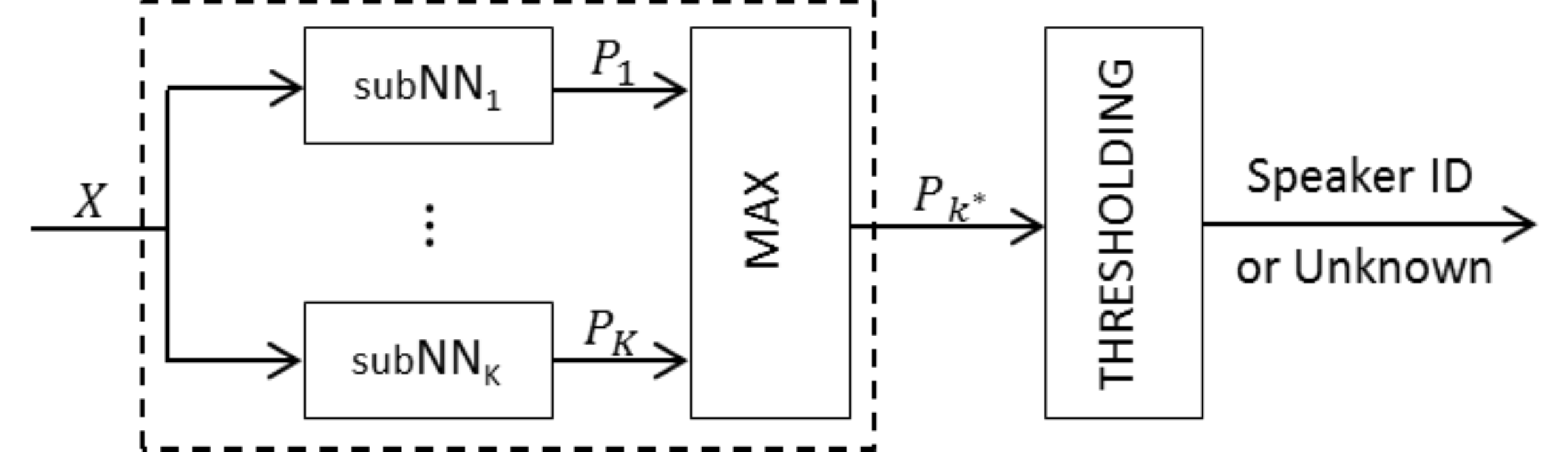}
	\caption{Open-set identification with $2$-class NN architecture.}
	\label{fig:subNN-Architecture}
\end{figure}

Solution based on $2$-class NN architecture is illustrated in Figure~\ref{fig:subNN-Architecture}. It consists of  K x $2$-class sub-networks (denoted as subNN in Figure~\ref{fig:subNN-Architecture}), which are trained in "one-against-all" configuration. This means that one NN per speaker is discriminatively trained to separate him/her form a set of background speakers. Instead of using directly features pooled from background speakers, the samples for the negative class are generated by the UBM on the fly. More details and discussion on the advantages of this approach are given in the next sections. The training results in a set of $K$ NN models $\{\lambda_k \}_{k=1}^{K}$. Each $\lambda_k$ is trained to output the posterior probability $p\left(\lambda_k | x_n \right)$, given the feature vector $x_n$ of the input audio utterance.

At the \textit{Closed-Set Identification Step}, "dashed box" in Figure~\ref{fig:subNN-Architecture}, the unknown audio stream is tested against all $K$ sub-networks. Further, assuming i.i.d. samples, the average log-posterior over the whole utterance is computed:
\begin{equation} \label{eq:post}
	P_k\left( X \right) = \frac{1}{N} \sum_{n=1}^{N} \log p\left(\lambda_k | x_n \right),
\end{equation}
and the largest posterior for the feature set $X$ is denoted as $P_{k^{*}}$

The $2$-class NN system \textit{Verification Step} is solved by thresholding the largest posterior probability speaker-independent threshold $\theta$:
\begin{equation} \label{eq:ver}
P_{k^{*}}\left( X \right) \begin{cases} 
  \geq\theta, & X \text{ is spoken by } s_{k^{*}} \\ 
  <\theta, & X \text{ is not spoken by } s_{k^{*}} 
  \end{cases}
\end{equation}

Even though there is no normalization with score from a UBM model, the thresholding of $P_{k^{*}}$ has similarities with thresholding $\Delta L$ as they are both measures of closeness of input feature vector set to particular speaker model, relative to the background model (simply because $1-P_{k^{*}}$ is the probability that $X$ belongs to the set of background speakers.).

\subsection{Multi-class NN architecture} \label{sec:multi-NN}
\begin{figure}[!thb]
	\centering
	\includegraphics[width=0.45\textwidth]{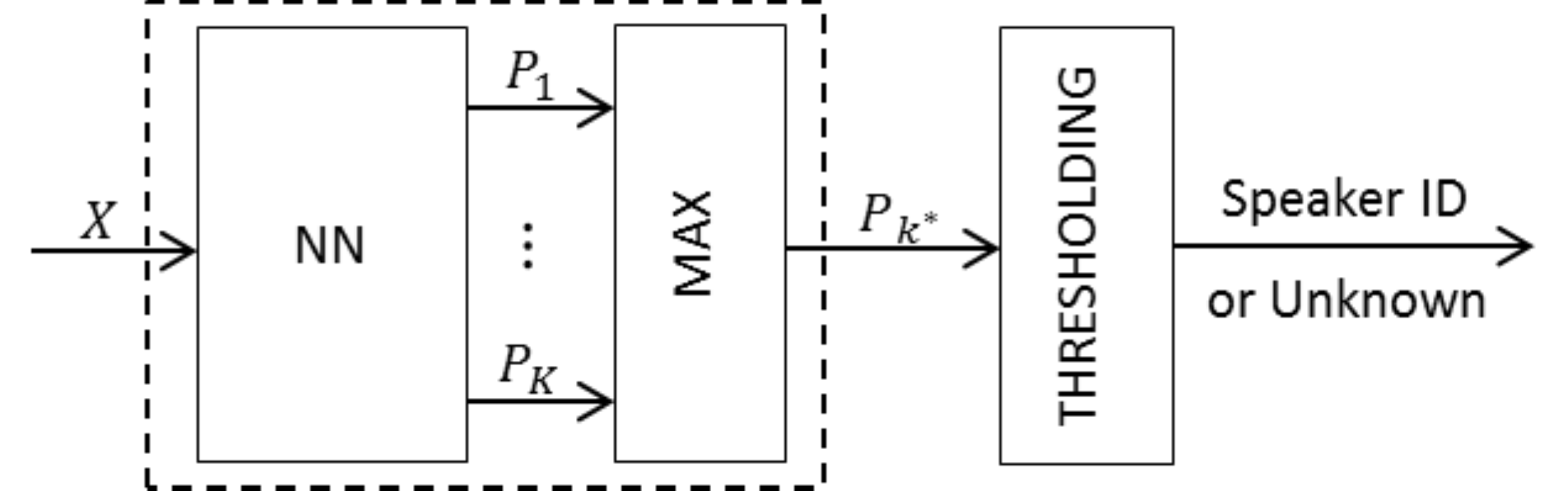}
	\caption{Multi-class NN architecture for open-set speaker identification scenario.}
	\label{fig:NN-Architecture}
\end{figure}

A common feature of previously discussed architectures is the training of a personalized model from a data set specific to a single person (in a pure generative way in Section~\ref{sec:GMM} and separating individual speaker from the background in Section~\ref{sec:subNN}). As a consequence such approaches cannot focus specifically on good discriminating features between individual speakers. This problem is addressed by a multi-output NN, see Figure~\ref{fig:NN-Architecture}, trained to discriminate between $K$ classes in a  \textit{all-together} configuration.

At \textit{Closed-Set Identification Step}, multi-class NN outputs $K$ posteriors, and the largest of the posterior probabilities $P_{k^{*}}$ is selected. Similarly to the system described in Section~\ref{sec:subNN}, 
the multi-class NN system \textit{Verification Step} is solved by thresholding the largest posterior probability, however this posterior probability has a different meaning. $P_{k^{*}}$, in Figure~\ref{fig:NN-Architecture}, is the probability that the input feature vector $X$ belongs to speaker $s_{k^{*}}$, however $1-P_{k^{*}}$ is not the probability that $X$ belongs to the set of background speakers, but the probability that $X$ belongs to the remaining set of enrolled speakers. This means that stability of thresholding $P_{k^{*}}$ depends on the size and distribution of the enrolled speaker set. For the simulations, reported in Section~\ref{sec:Experimetns} we use large population sizes of enrolled speakers, and hypothesize that the multi-class NN could learn the speakers space without a special "background speakers" class.

\subsection{Practical aspects and industrial applicability}
In this section we discuss some practical aspects of the presented architectures, which for real world applications, could be as important as the accuracy of these systems. A GMM based speaker classifier system has an important practical advantage in terms of modularity and extensibility. Most of the real-world speaker verification systems need to be extended with new speakers over time. Adding a new speaker to the set of enrolled speakers is rather simple when using GMMs. Once a UBM is obtained, it only requires the training of a speaker specific model for the new speaker. These nice properties are inherited by the $2$-class NN architecture, especially in the way it uses UBM to model average background speaker statistics. Both GMMs and $2$-class NNs do not need access to past audio data to enroll a new speaker. The reason is that sub-networks in the $2$-class NN system are trained to fit a single boundary between the target speaker data and average speech data statistics, sampled from the UBM. In contrast, enrolling a new speaker in a multi-class NN system, requires complete re-training and access to all audio data from all the speakers. On the other hand the multi-class NN system has an important advantage, which is low computational complexity at the recognition stage. Since the open-set speaker identification problem requires testing against all speaker models, the complexity of systems described in Section~\ref{sec:GMM} and \ref{sec:subNN}, explodes in applications with large population sizes. As an example, in our simulations with 700 enrolled speakers, a GMM system would need to test 701 models against $X$ and $2$-class NN system would need to test 700 models. Unless one has the possibility to completely parallelize the process, this is a significant disadvantage in comparison to the multi-class NN approach.

\section{Implementation} \label{sec:Implementation}
For each of the systems under test, different design choices were tried and parameter optimization performed to maximize the accuracy. Here we present the optimal configuration for each of the systems. For fair comparison the same feature extraction module was shared between all systems.

\subsection{Feature extraction} \label{sec:features}
In the first steps of the feature extraction module an energy-based Voice Activity Detector (VAD) is deployed to remove non-speech frames, and  a pre-emphasis filter $(1-\mu~z^{-1})$, with $\mu = 0.98$, is applied to the input speech. Next, Mel-Frequency Cepstral Coefficients (MFCCs) \cite{davis1980comparison} are extracted from a 20 ms anslysis window with 50\% overlap. Cepstral mean subtraction \cite{furui1981cepstralMean} is performed, and the resulting 24-dimensional feature vector $x_n$, corresponding to the $n$-th frame is ready to be used by the classification system.

\subsection{Implementation details of the GMM system} \label{sec:gmm_impl}
For the generative architecture, described in Section~\ref{sec:GMM}, individual speaker models $\omega_{k}$ were trained as GMMs with 64 components and diagonal covariance matrices. A UBM was trained with 1024 diagonal components. The same UBM was also used for the system described in Section~\ref{sec:subNN} to generate stochastic training samples with the desired distribution. The GMMs were fitted using the Expectation Maximization (EM) algorithm \cite{bilmes1998gentle}, initialized with k-means clustering \cite{Lloyd82GLA}.

\subsection{Implementation details of the $2$-class NN system}
For the $2$-class NN architecture, $K$ sub-networks were trained (one per speaker). Numerical values for the optimization parameters, used for training of the sub-networks, are listed in Table~\ref{tab:sub-NN-par}. Each sub-network is configured with an input layer of size 24, corresponding to a MFCC dimension, two hidden layers with 50 units each, and a single output. We used Rectified Linear Unit (ReLU) \cite{nair2010ReLU} activations for all hidden units and a softmax output layer to estimate the posterior probability for the particular target speakers. The network is trained with stochastic gradient descent to minimize the negative log-likelihood cost function. The global learning rate $\eta$ was set to $0.0001$. To speed up the training and optimize the convergence we used a variant of the momentum update, called Nesterov's Accelerated Gradient (NAG) \cite{nesterov1983method}, using the update equations described in \cite{bengio2013advances}, with parameter $\mu = 0.95$. RMS-prop \cite{tieleman2012lecture} was also used, with parameter for per-weight adaptive learning rate $\alpha = 0.99$. The number of epochs was set to $5$, and number the of samples per mini-batch to $800$.

\begin{table}[!thb] 
\caption{Training parameters for the $2$-class NN classifier.}
\label{tab:sub-NN-par}
  \centering
\begin{tabular}{lc} \toprule
    \textbf{Parameter}  & \textbf{Value} \\ \midrule
    Network architecture & [24in 50 50 1out] \\
    Learning rate $\eta$ & 0.0001 \\
    NAG $\mu$ & 0.95 \\
    RMS-prop $\alpha$ & 0.99 \\
    Number of epochs & 5 \\
    Batch size & 800 \\ \bottomrule
\end{tabular}
\end{table}

\subsection{Implementation details of the multi-class NN system}
For the multi-class NN architecture, one "all-together" network was trained. This single large network estimates the posterior probabilities for all $K$ target speakers. The architecture and training parameters of the multi-class NN are summarized in Table~\ref{tab:NN-par}. This network naturally has much wider hidden layers with 1200 units each, and $K$ output units. The  number of epochs in this case was set to $20$, and the number of samples per mini-batch to $15000$.

\begin{table}[!thb] 
\caption{Training parameters for the multi-class NN system. K in the "Network architecture" takes values \{100, 300, 500, 700\} based on the size of the enrollment set.}
\label{tab:NN-par}
  \centering
\begin{tabular}{lc} \toprule
    \textbf{Parameter}  & \textbf{Value} \\ \midrule
    Network architecture & [24in 1200 1200 \textit{K}out] \\
    Learning rate $\eta$ & 0.0001 \\
    NAG $\mu$ & 0.95 \\
    RMS-prop $\alpha$ & 0.99 \\
    Number of epochs & 20 \\
    Batch size & 15000 \\ \bottomrule 
\end{tabular}
\end{table}

\section{Experiments} \label{sec:Experimetns}
In this section we report results from an experiment with the three systems on enrolled sets with four different sizes \{100, 300, 500, 700\}, and an impostor set of 1400 speakers. The performance of the closed-set identification step ("dashed box" in Figures \ref{fig:GMM-Architecture}-\ref{fig:NN-Architecture})  is specified in terms of the Closed Set Recognition Rate (CSRR), which is the probability that $k^{*}$ is the same as the speaker ID of the test utterance. CSRR is calculated only over the enrolled speaker set. For the complete open-set speaker identification the Equal Error Rate (EER) is used to specify the performance.
In open-set speaker identification the misclassification of an utterance from an enrolled speaker is twofold. It can be either False Rejection or Mislabeling \cite{singer2004open_set}. For this situation we define the EER as the operating point where False Acceptance Rate equals the sum of False Rejection Rate and Mislabeling Rate.

\subsection{Data}
The dataset chosen for the experiments is the LibriSpeech ASR corpus, introduced in \cite{panayotov2015librispeech}, and used in the context of speaker recognition in  \cite{zeinali2016_libri_SRE}. It consists of 982 hours of wideband speech from 1281 male and 1202 female speakers. Every speaker has a different amount of recorded utterances and each utterance has a variable length between 3 and 30 seconds, with the average length of roughly 15 seconds. In the experiments outlined below, 70\% of the utterances were kept for the training set and the remaining 30\% as an independent test set. For the experiments, presented below, the entire data set of 2483 speakers is split into 3 non-overlapping sets: 383 speakers for training of the UBM, 1400 speakers to form the impostor set, and finally, 700 speakers for the enrolled set.

\subsection{Results} \label{sec:res}
CSRR, reported in Table~\ref{tab:recognition_rate_big} indicates that all systems have excellent performance in recognizing the correct speaker in the closed-set speaker identification step. When the population size increases, the performance of all systems deteriorate very slowly, and there is a clear trend that the multi-class NN begins to outperform the other solutions.
\begin{table}[!thb] 
	\caption{CSRR performance for the intermediate closed-set identification step with different sizes of the enrollment set.}
	\label{tab:recognition_rate_big}
	\centering
	\begin{tabular}{lcccc} 
		\toprule
                         & \multicolumn{4}{c}{\textbf{Population Size}} \\[0.1in]
    	\textbf{System}  & \textbf{100}     & \textbf{300}     & \textbf{500} & \textbf{700}	\\ \midrule     	
    	GMM         & \textbf{99.93 \%}	& 99.83 \%	& 99.80 \%	& 99.74 \%	\\
    	subNN   	 & 99.74 \%	& 99.70 \%   & 99.72 \%	& 99.67 \%	\\
    	NN          & \textbf{99.93 \%} & \textbf{99.92 \%} & \textbf{99.90 \%}   & \textbf{99.83 \%} \\
		\bottomrule
	\end{tabular}
\end{table}

The total performance of the tested systems is presented in terms of EER in Table~\ref{tab:eer_big}. The multi-class NN has an advantage for large population sizes, but the distance to the other systems diminishes quickly, and for the experiment with 100 enrolled speakers, the GMM based system performs best. The performance of all systems deteriorates rapidly with increase of population size (EER more than doubles from smallest to largest speaker set). The reason is that as the population size of enrolled speakers grows, it occupies more of the feature space and some impostor are incorrectly accepted, as they get  close to some of the enrolled speakers.

\begin{table}[!thb] 
	\caption{EER performance for open-set speaker identification with different sizes of the enrollment set.}
	\label{tab:eer_big}
	\centering
	\begin{tabular}{lcccc} 
		\toprule
                         & \multicolumn{4}{c}{\textbf{Population Size}} \\[0.1in]
    	\textbf{System}  & \textbf{100}     & \textbf{300}     & \textbf{500} & \textbf{700}	\\ \midrule    	
    	GMM         & \textbf{1.37 \%}	& 2.81 \%	& 3.13 \%	& 3.51 \%	\\
    	subNN   	 & 1.82 \%	& 3.07 \%   & 3.58 \%	& 4.04 \%	\\
    	NN          & 1.43 \% & \textbf{2.29 \%} & \textbf{2.65 \%}   & \textbf{3.04 \%} \\  
		\bottomrule
	\end{tabular}
\end{table}

\section{Discussion}
For large population sizes [300+] the multi-class NN system shows superior performance without any normalization of posterior probabilities. For small population sizes, as indicated in Table~\ref{tab:eer_big}, the advantage of multi-class system decreases, and perhaps extending the training with a class that models the background speaker space is required. Despite similar performance, there are other aspects of tested architectures that might determine their industrial use. $2$-class NN and GMM based classifiers are much more modular, and could be easily extended with additional enrolled speakers. However, in the open-set speaker identification scenario, these algorithms have high computational complexity at the recognition stage, which is not the case for the multi-class NN system.

\bibliographystyle{IEEEtran}

\end{document}